\documentclass{article}

\usepackage[final]{neurips_2020}

\usepackage[utf8]{inputenc} 
\usepackage[T1]{fontenc}    
\usepackage{hyperref}       
\usepackage{url}            
\usepackage{booktabs}       
\usepackage{amsfonts}       
\usepackage{nicefrac}       
\usepackage{microtype}      
\usepackage{kotex}
\usepackage{dsfont}
\usepackage{bm}
\usepackage{amsmath}
\usepackage{graphicx}
\usepackage{subfigure}
\usepackage{svg}

\usepackage{color}
\usepackage{soul}
\usepackage{lineno}

\newcommand{\jn}[1]{{\color{black}#1}}
\newcommand{\ours}{Ours}

\title{Tractable loss function and color image generation of multinary restricted Boltzmann machine}

\author{
Juno Hwang$^1$ \quad
Wonseok Hwang$^{2,*}$ \quad
Junghyo Jo$^{1,}$\thanks{Corresponding authors} \quad \\
$^1$Seoul National University, $^2$Clova AI, NAVER Corp.\\
\texttt{\{wnsdh10, jojunghyo\}@snu.ac.kr, wonseok.hwang@navercorp.com} \\
}

\begin{document}

\maketitle

\begin{abstract}
  The restricted Boltzmann machine (RBM) is a representative generative model based on the concept of statistical mechanics. In spite of the strong merit of interpretability, unavailability of backpropagation makes it less competitive  than other generative models. Here we derive differentiable loss functions for both binary and multinary RBMs. Then we demonstrate their learnability and performance by generating colored face images. 
\end{abstract}

\section{Introduction}
Generative models extract features from unlabeled data. They can infer underlying density of samples. Based on whether they use explicit or implicit density, two categories of generative models exist~\citep{Goodfellow2017}. Generative adversarial networks (GANs) are popular models with excellent performance using implicit density~\citep{Ian2014}. Restricted Boltzmann machines (RBMs) are representative models using explicit density~\citep{Hinton2012}. Since RBMs are developed with the basis of statistical mechanics, they provide strong interpretability with physics and information theory.
Nevertheless, their poor performance in many deep-learning tasks makes them less competitive.

RBMs are trained via Monte-Carlo sampling unlike other deep-learning models using gradient descent methods with loss functions. This limits RBMs to adopt modern neural architectures. For instance, deep belief networks(DBNs), consisting of multiple layers of RBMs, should be trained in a greedy layer-wise manner. 
Furthermore, once RBMs are extended to use multinary activation units, their computation becomes quickly intractable.  

Here we derive a loss function for RBMs to adopt gradient-based optimization with modern automatic differentiation libraries.  
Furthermore we generalize this idea applicable to multinary RBMs in which hidden states can take multiple values unlike the binary RBMs.
Then we demonstrate the learnability and performance of our model by generating images of MNIST, Fashion-MNIST, CIFAR-10, and CelebA. In particular, it is notable that the revised RBMs can generate colored images.

\section{Related works}
RBMs reproduce the underlying sample density of data by using neural networks with a pair of visible and hidden layers~\citep{Hinton2012}. Visible units are fully connected to hidden units, but units within the same layer have no connections to each other.
Given the graphical model, the probability of a sample $\mathbf{v}$ is defined as $p(\mathbf{v}) = \sum_{\mathbf{h}} p(\mathbf{v}, \mathbf{h})= \sum_{\mathbf{h}} Z^{-1} e^{-E(\mathbf{v}, \mathbf{h})}$ through the marginalization for hidden units $\mathbf{h}$ and the normalization with partition function $Z=\sum_{\mathbf{v, h}} e^{-E(\mathbf{v}, \mathbf{h})}$.
Here the RBM energy function is defined as
\begin{equation}
    E(\mathbf{v}, \mathbf{h}) = 
    - \sum_i a_i v_i - \sum_j b_j h_j - \sum_{i,j} v_i W_{ij} h_j
    = -\mathbf{a}\mathbf{v}^\top -\mathbf{b}\mathbf{h}^\top - \mathbf{v}\mathbf{W}\mathbf{h}^\top,
\end{equation}
where $v_i, h_j$ are the binary values of the $i$th visible and the $j$th hidden units, $\mathbf{a}, \mathbf{b}$ are their biases, and $W_{ij}$ is the weight between the $i$th visible unit and the $j$th hidden unit. 
We use bold symbols for representing row vectors and the capital bold symbol $\mathbf{W}$ for the weight matrix.

Original RBMs have binary hidden units. 
They have been modified to adopt multinary hidden units to increase the flexibility of internal representations of data.
One approach is to create $N$ copies of each hidden unit with the same weight and bias parameters~\citep{Teh2001}. Then the copied hidden units have a binomial distribution of the total activation value between 0 and $N$, while this model preserves the good mathematical properties of the binary RBM.
However, this structure has a problem that the variance of total activation depends on its mean value, making learning unstable. A small modification to the binomial units, called noisy rectified linear units (NReLU), solved this problem and worked better than binary units for several different tasks~\citep{Nair2010}.
Given infinite weight-sharing copies of binomial hidden units, NReLU considered equally-spaced offset biases such as $o_n = \{1/2, 3/2, 5/2, \cdots \}$ for each copy. 
The multinary RBM has a modified energy function of $E(\mathbf{v}, \mathbf{h}) = - \sum_i a_i v_i - \sum_j \sum_{n=1}^\infty (b_j - o_n) h_j - \sum_{i,j} v_i W_{ij} h_j$.
Then, compared to the expectation value $\mathbb{E}[h_j] = \sigma (x_j)$ of the hidden activity for the binary RBM, the mutinary RBM has a different mean expectation value:
\begin{equation}
    \mathbb{E}[h_j] = \sum_{n=1}^{\infty} \sigma(x_j-n+\frac{1}{2}) \approx \log (1+e^{x_j}) \approx \text{ ReLU}(x_j),
\end{equation}
where $\sigma(x) = (1 + e^{-x})^{-1}$ and $x_j=(\mathbf{b}+\mathbf{v}\mathbf{W})_j$.
Therefore, the mean activity can be interpreted as activity functions in general.
The multinary RBM should be trained still by Monte-Carlo sampling, which limits wider applications of RBMs. 

\begin{figure}[t]
\centering
\subfigure[]{
\includegraphics[scale=0.39]{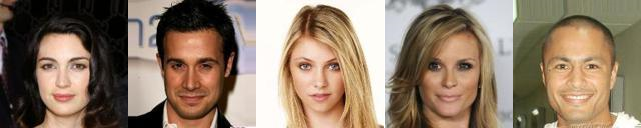}
}
\subfigure[]{
\includegraphics[scale=0.39]{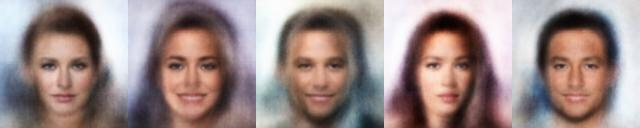}
}
\caption{
 (a) Cropped and resized 128×128 image samples from original CelebA dataset.
 (b) Generated samples from our model.
}
\end{figure}
\section{Model}
To make the RBM fully accessible with the powerful backpropagation algorithm, we derive a differentiable loss function for RBMs with binary hidden units, and generalize it for multinary hidden units. 
The original RBMs approximate the derivative of the likelihood of data. 
Here we approximate the likelihood itself.
\subsection{Loss function for binary RBMs}
Starting from the loglikelihood of a sample $\mathbf{v}$,
$\log p(\mathbf{v}) =  - \log Z + \log \sum_{\mathbf{h}} e^{- E(\mathbf{v}, \mathbf{h})}$, \citet{Hinton2012} defines a {\it free energy}:
\begin{equation}
\label{eq:free_energy}
    F(\mathbf{v}) = - \log \sum_{\mathbf{h}} \exp(\mathbf{a} \mathbf{v}^\top + \mathbf{b} \mathbf{h}^\top + \mathbf{v W}  \mathbf{h}^\top)
 = - \mathbf{a} \mathbf{v}^\top - \sum_{j=1}^H \log \big( 1 + \exp (\mathbf{b} + \mathbf{v W})_j \big),
\end{equation}
where $H$ numbers of binary hidden units are considered as $\mathbf{h} = (h_1, h_2, \cdots, h_H)$.
    
The partition function of $\log Z$ is usually intractable, but it can be approximated as $\log Z \approx - \sum_{\mathbf{v}'} F(\mathbf{v}')$, in which their gradients are approximated as the same in the exact sense.
Here $\mathbf{v}'$ is obtained from the Gibbs sampling of the Markov chain: $\mathbf{v} \rightarrow \mathbf{h} \rightarrow \mathbf{v}'$ where $\mathbb{E}[\mathbf{h}] = \sigma(\mathbf{b} + \mathbf{vW} )$ and $\mathbb{E}[\mathbf{v}'] = \sigma(\mathbf{a} + \mathbf{h} \mathbf{W}^\top)$.
This approximation has also been used in the contrastive divergence(CD) algorithm~\citep{Hinton2002}. Then the loss function $\mathcal{L}$ for entire data of $\{\mathbf{v}\}$ is defined as a contrastive free energy:
\begin{equation}
    \mathcal{L} \equiv \sum_{\{\mathbf{v}'\}} F (\mathbf{v}') - \sum_{\{\mathbf{v}\}} F (\mathbf{v}) = \sum_{\{(\mathbf{v}, \mathbf{v}')\}} \mathbf{a} (\mathbf{v}^\top - \mathbf{v}'^\top) + \sum_{j=1}^H \log \frac{1 + \exp (\mathbf{b} + \mathbf{v W})_j}{1 + \exp (\mathbf{b} + \mathbf{v}' \mathbf{W})_j}.
\end{equation}
Indeed, the derivative $\nabla \mathcal{L}$ of our loss function recovers the CD1 algorithm. 

\subsection{Loss function for multinary RBMs}
We observe that the binary RBM has a mean hidden activity of $\mathbb{E}[h_j] = \sigma(x_j)$ with $x_j = (\mathbf{b} + \mathbf{vW})_j$, and NReLU has $\mathbb{E}[h_j] = \sum_{n=1}^{\infty} \sigma(x_j - n + \frac{1}{2}) \approx \text{ReLU}(x_j)$.
Here $\text{ReLU}(x)$ activation function can be an approximation of the sum of sigmoid functions.

Extending this idea, any finite-valued activation function with its maximum $N$ can be approximated as the sum of sigmoid functions $f(x) = N g(kx) \approx \sum_{n=1}^N \sigma(x - o_n)$ with generalized offset biases $o_n$. $g(x)$ is an unit activation function and $k$ is a scale factor introduced for the mathematical convenience. $o_n$ can be determined from $o_n = f^{-1}(n - \frac{1}{2})$.
Note that the $\text{ReLU}(x)$ function has equally-spaced offset biases $o_n$.
The mutinary RBM has a mean hidden activity of $\mathbb{E}[h_j] = \sum_{n=1}^N \sigma(x_j - o_n) = f(x_j)$ and a variance of $\text{Var}[h_j] = \sum_{n=1}^N \sigma(x_j - o_n) \big( 1 - \sigma(x_j - o_n) \big) = f'(x_j) = Nk g'(kx_j)$, in which prime depicts derivative. 
Then, using Eq.~(\ref{eq:free_energy}), the data free energy for an $N$-multinary RBM with arbitrary offset bias $o_n$ is 
$F(\mathbf{v}) = - \mathbf{a} \mathbf{v}^\top - \sum_{j=1}^H \sum_{n=1}^N \log \big( 1 + \exp ((\mathbf{b} + \mathbf{v W})_j - o_n) \big)$.

Here we find an interesting equality
\begin{equation}
    \frac{d}{dx_j} \sum_{n=1}^N \log \big( 1 + \exp (x_j - o_n) \big) = \sum_{n=1}^N \sigma(x_j - o_n) \approx f(x_j). 
\end{equation}
Using this equality, the data free energy can be rewritten as $F(\mathbf{v}) \approx - \mathbf{a} \mathbf{v}^\top - \sum_{j=1}^H \int^{x_j} f(x) dx$.
Finally, this leads to the loss function of the multinary RBM as
\begin{equation}
\label{eq:mRBM}
    \mathcal{L} \equiv \sum_{\{(\mathbf{v}, \mathbf{v}')\}} \mathbf{a} (\mathbf{v}^\top - \mathbf{v}'^\top) + \sum_{j=1}^H \int_{(\mathbf{b}+\mathbf{v}'\mathbf{W})_j}^{(\mathbf{b}+\mathbf{v}\mathbf{W})_j} f(x) dx,
\end{equation}
where the expectation of reconstructed visible activity is given by $\mathbb{E}[\mathbf{v}'] = f(\mathbf{a} + f(\mathbf{b} + \mathbf{vW}) \mathbf{W}^\top)$.

\section{Experiments}
For a demonstration of the learnability of our model, we adopt a sigmoid function as our unit activation function of $g(x) = \sigma(x)$.
Our resulting multinary RBM has a general activation function $f(x) = N \sigma(kx)$, and a loss function of $\mathcal{L} = \sum \mathbf{a}(\mathbf{v}^\top-\mathbf{v'}^\top) + \frac{N}{k}\sum_{j=1}^H \log \frac{1+\exp (k(\mathbf{b+vW}))} {1+\exp (k(\mathbf{b+v'W}))}$ from Eq.~(\ref{eq:mRBM}).
To generate color images, we set $N=255$ for both visible and hidden nodes, and adopt convolutional deep belief networks (CDBNs)~\citep{Lee2009} with following modification.
Since it is difficult to use probabilistic max pooling for multinary units, we use strided convolution layer instead of pooling layer. In the generation task, we stack five convolutional RBMs (channel sizes: 3--64--128--256--256, stride: 2, filter size: 4×4) followed by four fully-connected RBMs that gradually decrease the feature dimension (4096--1000--500--200--100). In the classification task, fully-connected layers are removed and the number of channels are adjusted depending on datasets.
The model is trained via backpropagation algorithm after summing all losses from individual layers. The batch size is set to 128. ADAM optimizer is used with learning rate 0.03, and decay rates $\beta_1$ = 0.9, $\beta_2$ = 0.999. The training stops after 500 epochs. In case of CelebA 128x128 generation task, it takes approximately two days on single NVIDIA P40 GPU.


\begin{figure}[t]
    \centering
    \includegraphics[scale=0.2]{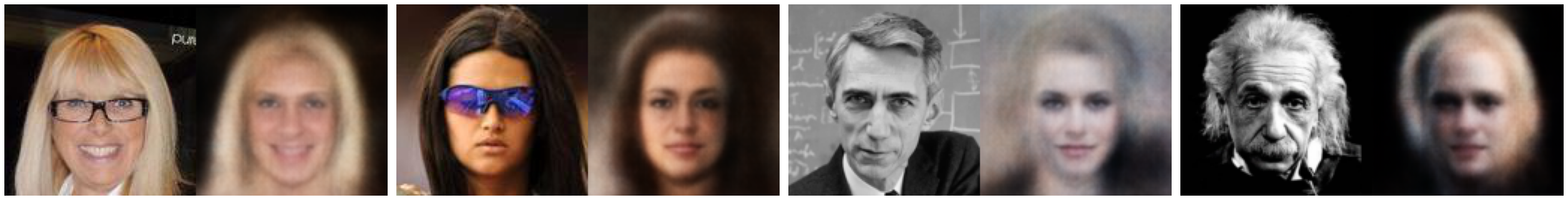}
    \caption{Original images and their reconstructions. The two on the left are from the CelebA dataset, and the two on the right are not.}
\end{figure}

{\bf Quantitative results.}
\begingroup
\setlength{\tabcolsep}{2.4pt} 
\renewcommand{\arraystretch}{0.9} 
\begin{table}[t!]
\centering
\small
\begin{tabular}{lrrrr}
\toprule
Model   & MNIST & Fashion-MNIST & CIFAR-10 & Caltech-101 \\ \midrule

\citet{Lee2009}    &      \textbf{0.82\%}&           -  &               -  &            34.6\% \\

\citet{ranzato2010modeling}  &  - &-  &    \textbf{29.0\%}  &   - \\

\citet{goh2013top} &   0.91\% &  - & - & \textbf{20.3\%} \\

\citet{zhang2019deep}     & 1.94\%  & 12.31\% & 44.37\%  & - \\
\midrule
\ours ($N=1$)     & 2.44\%  & 13.12\%  & 58.72\%  & 71.02\% \\
\ours ($N=255$)     & 1.23\%  & \textbf{9.26\%}  & 35.78\%  & 34.74\% \\   
\bottomrule\\
\end{tabular}
\caption{Test errors of various datasets. For the Caltech-101 dataset, we randomly selected 30 training samples for each category to train the model.}
\vspace{-4mm}
\end{table}
\endgroup
\begingroup
\setlength{\tabcolsep}{2.4pt} 
\renewcommand{\arraystretch}{0.9} 
\begin{table}[t!]
\centering
\small
\begin{tabular}{lrr}
\toprule
Model   & {CelebA 64×64} & {CelebA 128×128} \\ \midrule
HDCGAN \citep{curto2017high}  & 8.44                             & -                                    \\
PAE \citep{bohm2020probabilistic}      & 49.2                             & -                                    \\
COCO-GAN \citep{lin2019coco} & -                                & \textbf{5.74}                               \\
Pres-GAN \citep{dieng2019prescribed} & -                                & 29.115                             \\ 
AGE \citep{heljakka2018pioneer}& 26.53                                & 154.79                             \\
Pioneer \citep{heljakka2018pioneer}& \textbf{8.09}                                & 23.15                             \\
\ours     & 101.05                           & 106.03                          \\   \bottomrule\\
\end{tabular}
\caption{The FID scores (lower is better)}
\vspace{-6mm}
\end{table}
\endgroup
We evaluate our models on two tasks: (i) image classification and (ii) image generations. 
For the first task, the hidden units from final layers are fed into a support vector machine.
Table 1 summarizes the results showing that our model can faithfully extract features from various types of images. Notably, our model achieves the best result for Fashion-MNIST among RBM-based models. \jn{Overall, however, it did not stand out in classification task. We think the absence of the pooling layer has degraded classification performance.}
For the generation task, we use CelebA dataset \jn{with the aim of facilitating qualitative analysis,} that is cropped and resized into  64×64 and 128×128 pixels. Our model can produce a variety of skin colors, facial expressions, and gender (Fig. 1). The image reconstruction experiments also show the interesting generalization ability of our model such as the removal of facial accessory (left two examples in Fig. 2) and the colorization of the gray images (right two examples in Fig. 2).
Notably, as far as we know, our model is the only model that can produce high-quality face images in color using pure RBMs.
Table 2 shows Fréchet inception distance(FID) score for quantitative comparison.  



{\bf Ablation study.}
The bottom two rows of Table 1 show a comparison of the classification performance of differentiable binary($N=1$) and multinary($N=255$) DBNs with the same hyperparameters and network structure. In all cases, multinary units showed improved performance. The result is consistent with previous research~\citep{Nair2010} showing that the performance of RBMs is improved since the ReLU hidden unit is more expressive than binary units.

\section{Conclusion}
In summary, we derived differentiable loss functions for binary and multinary RBMs with arbitrary activation functions, and demonstrated its performance by generating color images.
Based on our knowledge, this is the first RBM-based model that generates colored face images.
%

This study shows the prospect of RBMs as a machine for generating high-resolution images. The multinary RBMs with differentiable loss functions have ample rooms for improvement. Although we used a simple sigmoid function, our model is designed to consider any arbitrary activation function. Moreover, it is expected that RBMs will be much more competitive in image generation tasks with further theoretical developments such as the implementation of pooling layers and pseudo-likelihood of entire stacked RBMs.


\newpage
\small
\bibliography{references}
\bibliographystyle{acl_natbib}
\medskip




\clearpage

\setcounter{figure}{0}
\renewcommand{\thefigure}{A\arabic{figure}}
\appendix
\normalsize
\section{Appendix}

\begin{figure}[h]
    \centering
    \subfigure[]{
        \includegraphics[scale=0.35]{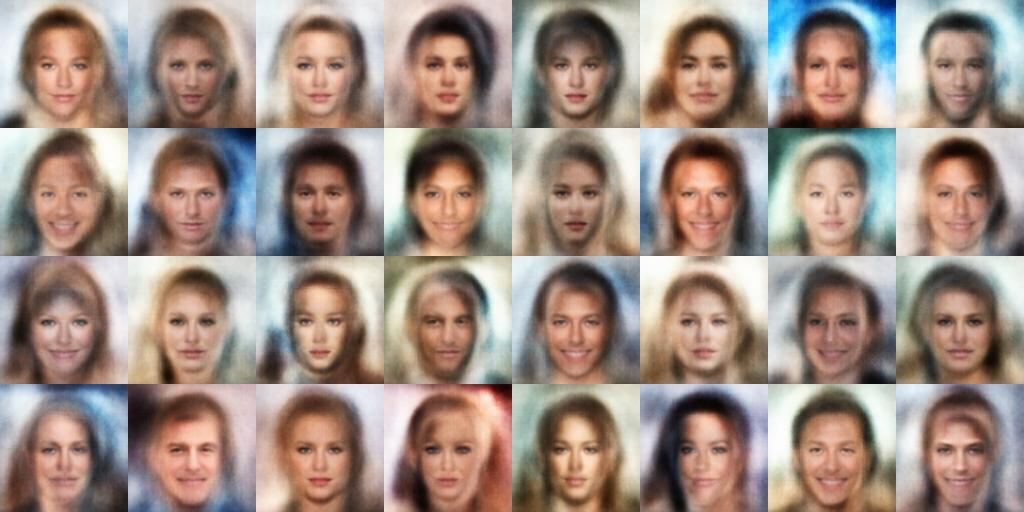}
    }
    \subfigure[]{
        \includegraphics[scale=0.35]{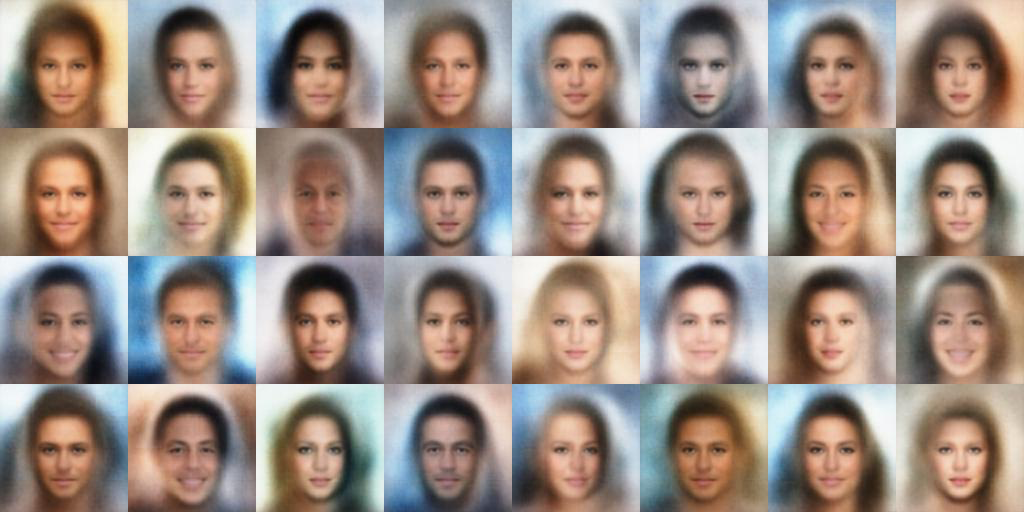}
    }
    \caption{
     (a) Samples generated from the model in the text.
     (b) Samples generated from the model with the same structure and conditions, but with a reduced learning rate (1/3 times).
}
\end{figure}

  We show more images generated from the model after training CelebA dataset (Fig. A1). 
  Here we compare images generated by using two different learning rates. When we used 1/3 times smaller learning rate (Fig. A1b), the generated face images look more blurred and averaged, although the key features of the face remain clear. At this stage, we do not understand how the learning rate affects the quality of generated images. 
  
  
  Our reconstruction removed eyeglasses from original face images (Fig. 2). This may result from the averaging effect of training samples. Thus we checked that only 4.67\% (9,459 out of the total 202,599) of the samples in CelebA dataset are tagged for eyeglasses according to the annotation attributes provided~\citep{liu2015faceattributes}.

\end{document}